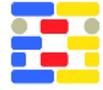

# CROWDSOURCED-BASED DEEP CONVOLUTIONAL NETWORKS FOR URBAN FLOOD DEPTH MAPPING

Bahareh Alizadeh[1], Amir H. Behzadan[1]
[1]Texas A&M University, College Station, Texas, U.S.

## Abstract

Successful flood recovery and evacuation require access to reliable flood depth information. Most existing flood mapping tools do not provide real-time flood maps of inundated streets in and around residential areas. In this paper, a deep convolutional network is used to determine flood depth with high spatial resolution by analyzing crowdsourced images of submerged traffic signs. Testing the model on photos from a recent flood in the U.S. and Canada yields a mean absolute error of 6.978 in., which is on par with previous studies, thus demonstrating the applicability of this approach to low-cost, accurate, and real-time flood risk mapping.

## Introduction

Flood patterns have been altered in many parts of the world due to the change in climate and the resulting change in hydrological cycles (Bowes et al., 2021; Alfieri et al 2017; Ward et al 2014; Arnell and Gosling 2016). Between 2015 and 2020, an annual average of 16.2 climate disasters were reported in the United States (Smith, 2021). In November 2021, for example, a major rainfall event spurred a series of floods in northern parts of the U.S. state of Washington and southern parts of the Canadian province of British Columbia. In British Colombia, this flood caused record-breaking insured damage of $450 million (Insurance Bureau of Canada, 2021). Moreover, at least 500 people were displaced, as estimated by the Center for Western Weather and Water Extremes (CW3E) (2021).

The current practice of flood mapping involves logging the variation in flood depth at the location of pre-installed stream gauges primarily in riverine areas. However, it is equally important to identify and map flood-inundated roads in and around residential areas because disruptions in the road network can severely hamper access to shelters and the movement of goods and services during and after a flood event (Arabi et al., 2021), as well as compromise the safe evacuation of residents and motorists (Gebrehiwot et al., 2019). A significant number of flood-related deaths belong to motorists who attempt to cross inundated roads while underestimating flood depth and water velocity (Federal Emergency Management Agency, 2007; Sharif et al., 2012). Data from the U.S. state of Texas between 1959 and 2019, for example, reveals that there were 570 vehicle-related flood fatalities during this period, accounting for 58% of total flood fatalities (Han and Sharif, 2020). In Texas alone, it is estimated that 93% of flood-related fatalities occur as a result of walking or driving through flooded roads (Sharif et al., 2015).

Flood risk management has gained increasing interest in recent decades due to the emergence of advanced technologies including light detection and ranging (LiDAR), synthetic-aperture radar (SAR), and satellite imagery (Mishra et al., 2022). However, the high operational cost, adverse weather conditions, and variations in surface topography limit the application of these technologies (Stone et al., 2000; Cross et al., 2020; Forati and Ghose, 2022; Kamari and Ham, 2021). In addition, most such tools and resulting data are not readily available to the public during a flood event. Current public resources for acquiring immediate flood depth information use data transmitted by stream gauges near rivers and coastal areas (Lo et al., 2015). These gauges need to be frequently calibrated and maintained. Moreover, due to their limited number, resulting flood depth data lacks spatial granularity sufficient for decision-making along local evacuation roads and in residential areas.

Recent advancements in artificial intelligence (AI) and image processing have provided new opportunities for estimating flood depth from user-contributed photos. In this research, photos depicting submerged stop signs are of interest because stop signs (and in general, traffic signs) have standardized shapes, colors, and sizes, and are omnipresent in many residential neighborhoods, making them ideal measurement benchmarks. A deep learning model is utilized to detect stop signs in street photos and measure the length of the visible part of the submerged pole. Next, flood depth at the location of the stop sign is calculated by comparing the pole length before and after the flood. Using this approach, users (i.e., residents, search and rescue teams) can rapidly estimate water levels in their surroundings, and use this information to identify safe transit routes. The method is generalizable to a wide range of stop sign images regardless of the geographical region, as long as the shape, color, and dimensions of these signs are standardized by corresponding traffic authorities.

## Methods of implementation

According to the Manual on Uniform and Traffic Control Devices (MUTCD) (Federal Highway Administration, 2004), regulatory stop signs in the U.S. have standardized width and height of 30 inches on single-lane roads and 36 inches on multi-lane roads and expressways. The scope of the research presented in this paper is residential areas, and therefore, stop signs with 30 inches height and width are of interest. Due to their standardized size, stop signs can be used as a benchmark for measuring flood depth if both pre- and post-flood photos of the same stop sign are available during analysis. In a nutshell, given the height of a stop sign in an image measured in pixels ($S$) and in inches, a ratio of pixels to inches can be calculated, and subsequently applied to the pole length in pixels ($P$) to calculate the same length in inches (Figure 1). By subtracting the visible parts of the pole in pre- and post-flood photos, flood depth is approximated at the location of the stop sign.

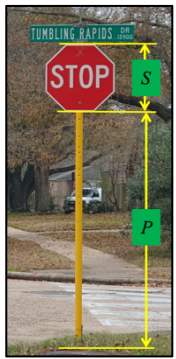

*Figure 1: Pole length estimation by comparing the number of pixels corresponding to sign and pole components*

For visual recognition of stop signs and their poles in pre- and post-flood photos, the You-Only-Look-Once (YOLO) model is utilized which is a fast object detector that predicts bounding boxes and class probabilities in one round of image evaluation using a single neural network (Redmon et al., 2016). Compared to other object detection models, YOLO can perform in real-time, as measured by the number of frames per second (FPS), with relatively high accuracy and less computation cost. For example, RetinaNet-101-500 (Lin et al., 2017), R-FCN, SSD321 (Liu et al., 2016), and DSSD321 (Fu et al., 2017) yield mean average precision (mAP) of 53.1% (at 11 FPS), 51.9% (at 12 FPS), 45.4% (at 16 FPS), and 46.1% (at 12 FPS) on the Microsoft COCO dataset (Lin et al., 2014), respectively. The YOLOv4 proposed by Bochkovskiy et al. (2020) is the latest official version of the model and can achieve a mean average precision (mAP) of 65.7% at 65 FPS on the Microsoft COCO dataset. In this research, the pretrained YOLOv4 model on the Microsoft COCO dataset is initially selected which can detect 80 object classes including stop signs. Using transfer learning (Han et al., 2018; Tammina, 2019), all pre-trained weights are kept unchanged, except for the last three YOLO layers which are trained on an in-house dataset of annotated stop signs and poles.

The architecture of the YOLOv4 model is illustrated in Figure 2. The cross-stage-partial-connections (CSP) network along with the Darknet-53 (Wang et al., 2020) serve as the backbone of the model for extracting essential features from the input image. The neck comprises a spatial pyramid pooling network (SPP) (He et al., 2015) that runs from bottom to top, as well as a path aggregation network (PANet) (Liu et al., 2018) that runs from top to bottom. The neck enables the collection of feature maps from the last three stages of the backbone. The head of the YOLOv4 model is similar to that of the YOLOv3 model which outputs the coordinates, and the width and height of predicted bounding boxes using the feature pyramid network (FPN) (Lin et al., 2017). The input of the model is a single image with the resolution of $320 \times 320 \times 3$ (height × width × RGB channel), while the output consists of predicted bounding boxes over two classes, namely stop sign and pole.

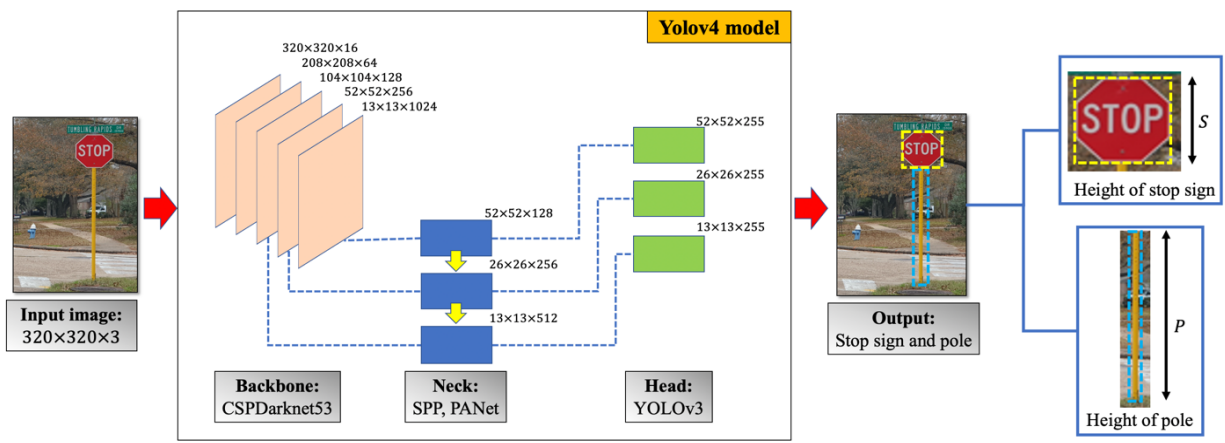

*Figure 2: Architecture of the YOLOv4 model*

## Training data description

An in-house dataset is created for model training including pre-flood and post-flood photos of stop signs in different languages (e.g., English, Spanish, French, and Arabic). The training set for pre-flood photos consists of 334 photos of stop signs extracted from the Microsoft COCO dataset excluding images in which either the sign or pole is not fully visible. The training set for post-flood photos includes 270 web-mined photos. Web-mining is conducted in social media using relevant keywords such as "flood stop sign" and their translations in other languages. Stop signs and their poles are subsequently annotated in all photos. Training the deep learning model on stop signs in various forms and languages may result in learning overly detailed features potentially leading to false detection of other traffic signs that are visually proximal to stop signs. To resolve this problem, the training set is further enriched with images of other traffic signs with no annotations (zero label). In particular, 71 web-mined post-flood photos and 61 pre-flood photos of zero label are added to the training set, to allow the model to learn distinctive features specific to stop signs while avoiding detecting other traffic signs.

To balance the training set with respect to the number of pre- and post-flood photos, synthetic data augmentation is conducted by manually generating synthetic images. Synthetic image generation has been widely used as a method of data augmentation in past research (Hu et al., 2021; Tremblay et al., 2018; Shaghaghian and Yan, 2019). For the post-flood subset, 64 synthetic photos are generated using photo editing tools by collecting post-flood photos of other traffic signs and replacing the signs with stop signs. Newly added synthetic data are annotated in the same way as annotating real images of stop signs.

## Model development

The model is trained on the in-house training set over 4,000 iterations using a learning rate of 0.001 with a batch size of 1 and subdivision of 64. The Adaptive Moment Estimation (Adam) method (Kingma and Ba, 2014) is used to accelerate model optimization by choosing a separate learning rate for each parameter (Ruder, 2016). As a validated algorithm for first-order gradient-based optimization of stochastic objective functions, Adam Optimizer works best for sparse input data and has low computation cost and high productivity (Sharma et al., 2019). The backbone of the object detection model (Darknet-53) is built in Windows on a Lenovo ThinkPad laptop computer with 7 cores, 9750H CPU, 16 GB RAM, and Nvidia Quadro T1000 GPU with a 4 GB memory. Network resolution (i.e., image input size) is set to 320×320×3 to reduce computational cost and time.

In addition to synthetic data augmentation previously described, random and real-time data augmentation on the training set is also enabled in the YOLOv4 model to increase the size of the training set by creating slightly modified copies of training images. In particular, hue, saturation, and exposure of training samples are modified within [-18…+18], [0.66…1.5], and [0.66…1.5], respectively as recommended by Bochkovskiy et al. (2020). In addition, half of the training images are flipped horizontally, but no image is flipped vertically as recommended by Hu et al. (2020). Mosaic data augmentation is also applied to 50% of images by combining groups of four different images to create a new image (Hao and Zhili, 2020). The total processing time for training the model on the in-house dataset is approximately 12 hours, with an average loss of 0.567.

## Model validation

Five-fold cross validation (Lyons et al., 2018; Fetanat et al., 2021) is implemented to validate the model and avoid overfitting on the training set. Using this approach, the training set is randomly split into two subsets (containing 80% and 20% of the training data, respectively) for five times. The model is then trained on the larger subset and validated on the smaller subset for five rounds, and mAP is measured as the aggregate of class-specific average precision (AP) values (Turpin and Scholer, 2006). The AP for each class is calculated using intersection over union (IoU), which computes the overlap between predicted bounding boxes and ground-truth bounding boxes (Nath and Behzadan, 2020). If IoU is above a predefined threshold (typically 50%), the detection is marked as correct; otherwise, it is marked as incorrect (Zhu et al., 2021). The optimum number of iterations is determined as the average of the optimum number of iterations (corresponding to the highest mAP) for each validation set. Five-fold cross validation yields an average IoU of 82.26%, AP of 97.37% and 96.70% over the two classes (stop sign and pole), and mAP and average iteration numbers of 97.04% and ~3,000. The model is then re-trained one last time with the optimum iteration number to ensure that it has not overfitted on the training set, and is sufficiently generalizable to the unseen test set.

## Techniques to overcome challenging cases

Model training on street view images poses unique challenges that need to be addressed to improve the outcome. One such issue occurs when the model detects more than one stop sign in the input image, with one appearing closer to the viewpoint and other(s) at a farther distance, as shown in Figure 3(a). While the object detection model is robust enough to detect multiple stop signs, the problem arises since each photo bears only a single geolocation tag which, by default, is used to map the detected stop sign. In the presence of multiple detected stop signs, however, this one-to-one correspondence will no longer be valid. To remedy this problem, when multiple stop signs are detected, the largest stop sign (appearing closest to the camera viewpoint) is selected as the target output. Among detected poles, the one with the closest center of *x* coordinates to the center of *x* coordinates of the candidate stop sign is then selected as the target pole.

Another challenge is reflection in water, which is a common problem in object detection. Simple solutions for reflection removal are based on the analysis of multiple input images, including pairs of images that are taken from different orientations, or with different polarizations (Sarel and Irani, 2004; Kong et al., 2013). By comparison, single-image reflection removal is more challenging. Wan et al. (2016) proposed a multi-scale depth of field (DoF) strategy to classify edge pixels and remove reflection in the image background. Fan et al. (2017) addressed the reflection issue using a deep learning model that extracts edge information to identify low-level vision tasks, and demonstrated high quality layer separation. Specifically in the post-flood image dataset, there are several cases where the stop sign is reflected in water, as shown in Figure 3(b). To remedy this, the strategy is to not label the reflected stop sign and pole, thus forcing the model to learn from the difference between color intensity of pixels in actual and reflected objects.

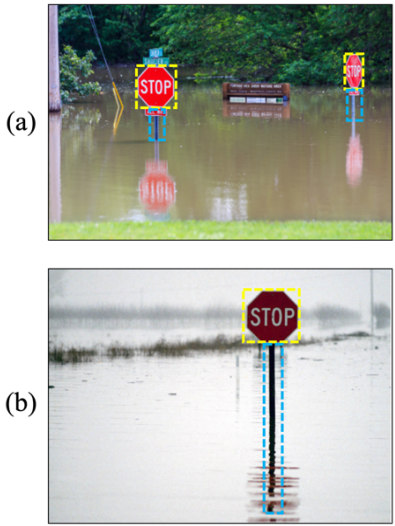

*Figure 3: Demonstration of challenging cases in post-flood photos: (a) multiple detections (base photo courtesy of Steve Zumwalt/FEMA), and (b) reflection in water (base photo courtesy of TSGT Mike Moore, USAF)*

**Model performance on test data**

To quantify model performance, a test set containing crowdsourced images of the 2021 flood in the Pacific Northwest was collected. During this flood event, the U.S. Geological Survey (USGS) stream gauge at Nooksack River in Ferndale, WA reported a peak stage height of 150.35 ft., indicating an increase of 5 ft. in water level (National Oceanic and Atmospheric Administration, 2021). The flood stage at the USGS Skagit River stream gauge near Mt. Vernon, WA was also rising above the major flood stage (32 ft.), logging 37 ft. of floodwater at this location (Center for Western Weather and Water Extremes, 2021), before being destroyed by flood debris (National Oceanic and Atmospheric Administration, 2021). In the following section, the ability of the proposed model is to estimate flood depth using stop signs as measurement benchmarks is demonstrated. To create the test set, 11 crowdsourced photos containing 3 stop signs in the U.S. state of Washington and 8 stop signs in the Canadian province of British Colombia are web-mined. The locations of these stop signs are shown on the map in Figure 4. Each of the 11 post-flood images is first paired with its corresponding pre-flood photo (depicting the same stop sign), and then annotated with two classes of stop sign and pole.

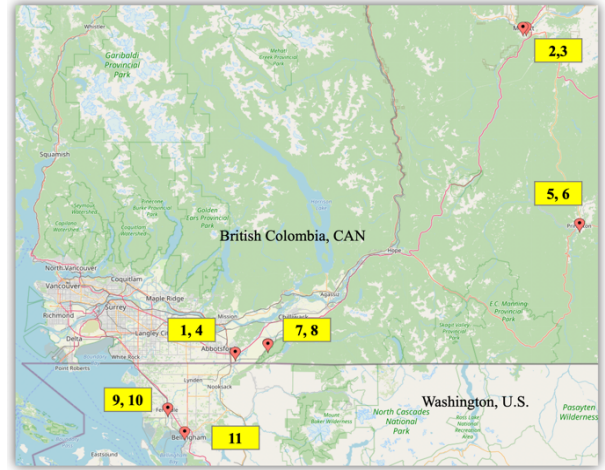

*Figure 4: Map of 11 test data related to the 2021 Pacific Northwest flood (numbers refer to IDs shown in Table 1)*

A commonly used metric for describing the discrepancy in flood depth estimation is mean absolute error (MAE) (Chaudhary et al., 2019; Cohen et al., 2019; Park et al., 2021; Alizadeh and Behzadan, 2020; Alizadeh et al., 2021). As shown in Equation (1), the MAE for pole length estimation ($MAE_P$) can be calculated as the average of the absolute difference between detected pole length ($P'$) and ground-truth pole length ($P$) over $N$ photos (including pre- and post-flood photos). The model obtains $MAE_P$ of 3.916 in. and 6.769 in. for pole length estimation in pre- and post-flood photos of the test set, respectively. The MAE for flood depth estimation ($MAE_D$) can be calculated using Equation (2) as the average of the sum of the absolute values of the difference between ground-truth pole length ($P$) and detected pole length ($P'$) in pre- and post-flood photos over $M$ pairs of photos. Table 1 shows the $MAE_D$ achieved by the model for flood depth estimation for each photo in the test set. The MAE of flood depth estimation is 6.978 in., which is significantly improved compared to the MAE of 12.65 in. reported previously by Alizadeh and Behzadan (2021) and Alizadeh et al. (2021) through processing images of submerged stop signs. Also, this new MAE is on par with other studies on flood depth mapping using image processing techniques. For example, Cohen et al. (2019) reported an MAE of 7-12 in. in coastal and riverine areas. Chaudhary et al. (2019) obtained an MAE of 4 in. by detecting submerged objects in social media images, and comparing them with their predefined sizes. Park et al. (2021) reported an MAE of 2.5 in. by comparing images of submerged vehicles with their predefined sizes.

$$MAE_P = \frac{1}{N}\sum_1^N |P' - P| \quad (1)$$

$$MAE_D = \frac{1}{M}\sum_1^M \left(|P'_{pre} - P_{pre}| + |P'_{post} - P_{post}|\right) \quad (2)$$

While MAE can be used as one descriptor of the reliability of the model, it must be noted that the main benefit of using the approach presented in this paper is the high granularity (spatial resolution) of datapoints for flood depth mapping considering the large number of stop signs mounted in various roads and streets in residential areas.

*Table 1: Performance of the model on flood depth estimation*

| ID | Location | Flood depth | | |
| --- | --- | --- | --- | --- |
| | | Detection (in.) | Ground-truth (in.) | Δ (in.) |
| 1 | CAN/Abbotsford | 25.486 | 27.162 | -1.676 |
| 2 | CAN/Merritt | 13.949 | 9.759 | 4.190 |
| 3 | CAN/Merritt | 18.722 | 13.436 | 5.286 |
| 4 | CAN/Abbotsford | 20.292 | 20.157 | 0.135 |
| 5 | CAN/Princeton | 0.463 | 6.844 | -6.381 |
| 6 | CAN/Princeton | 51.243 | 15.780 | 35.463 |
| 7 | CAN/Abbotsford | 4.871 | 3.968 | 0.903 |
| 8 | CAN/Abbotsford | 9.331 | 6.679 | 2.652 |
| 9 | USA/Ferndale | 44.342 | 41.342 | 3.000 |
| 10 | USA/Ferndale | 26.217 | 10.923 | 15.295 |
| 11 | USA/Bellingham | 9.820 | 8.038 | 1.782 |
| | $MAE_D$ | | | 6.978 |

## Conclusions

The scientific community agrees that climate change is leading to more severe flooding events in various parts of the world (Alfieri et al., 2017; Ward et al., 2014; Arnell and Gosling, 2016). Despite the importance of reliable flood depth data to many stages of flood mitigation (ranging from evacuation to recovery), most existing flood mapping techniques only estimate flood inundation over large areas (e.g., regional) leaving it primarily to the decision-maker to interpret and/or extrapolate results for door-to-door applications that require highly granular (i.e., street-level) flood depth data. In this paper, a deep neural network, namely YOLOv4, was adopted using the transfer learning technique on an in-house dataset of crowdsourced stop signs photos. The model was trained on pre-flood (334 photos from the Microsoft COCO dataset, and 61 web-mined photos of other traffic signs with zero label) and post-flood (270 web-mined photos, 71 web-mined photos of other traffic signs with zero label, and 64 synthetic photos) photos of stop signs. To address the overfitting problem on the training set, the optimum number of iterations was obtained using 5-fold cross validation. The trained model was then used to estimate flood depth at the location of the stop sign, by comparing the visible parts of the pole length in pre- and post-flood photos. Testing the model on 11 sample photos from the 2021 Pacific Northwest flood yielded an MAE of 6.978 in. for flood depth estimation which is on par with error values reported in previous studies. Since stop signs have standardized shapes, colors, and sizes, and are mounted in many road intersections, this approach can be generalized to any geographical location with regulated and standardized stop signs to help estimate flood depth at the street level, providing that enough crowdsourced data is available. In summary, the main contributions of this research are: (1) an in-house image dataset of submerged stop signs which was used for training, validation, and testing of the deep convolutional networks; (2) the generalizability of the proposed method to other regions since traffic signs (an in particular, stop signs) are omnipresent and easily recognizable in many parts of the world; and (3) the ability to generate high spatial resolution flood maps by significantly increasing the number of datapoints where floodwater depth estimation can be calculated.

## Acknowledgments

This study is funded by award #NA18OAR4170088 from the National Oceanic and Atmospheric Administration (NOAA), U.S. Department of Commerce. The authors would also like to thank Mr. Nathan Young for his assistance in data collection. Any opinions, findings, conclusions, and recommendations expressed in this paper are those of the authors and do not necessarily represent the views of the NOAA, Department of Commerce, or the individual named above.

# References

Alfieri, L., Bisselink, B., Dottori, F., Naumann, G., de Roo, A., Salamon, P., Wyser, K., & Feyen, L. (2017) Global projections of river flood risk in a warmer world. Earth's Future, 5(2), p.pp. 171-182. https://doi.org/10.1002/2016EF000485

Alizadeh Kharazi, B., & Behzadan, A. H. (2021) Flood depth mapping in street photos with image processing and deep neural networks. Computers, Environment and Urban Systems, 88, p.pp. 101628. https://doi.org/10.1016/j.compenvurbsys.2021.101628

Alizadeh, B., Li, D., Zhang, Z., & Behzadan, A. H. (2021) Feasibility study of urban flood mapping using traffic signs for route optimization. In the proceeding of 28th EG-ICE International Workshop on Intelligent Computing in Engineering, p.pp. 572-581, Berlin, Germany. https://arxiv.org/abs/2109.11712

Arabi, M., Hyun, K., & Mattingly, S. P. (2021) Adaptable resilience assessment framework to evaluate an impact of a disruptive event on freight operations. Transportation Research Record, 2675(12), p.pp. 1327-1344. https://doi.org/10.1177/03611981211033864

Arnell, N. W., & Gosling, S. N. (2016) The impacts of climate change on river flood risk at the global scale. Climatic Change, 134(3), p.pp. 387-401. https://doi.org/10.1007/s10584-014-1084-5

Bochkovskiy, A., Wang, C. Y., & Liao, H. Y. M. (2020) Yolov4: Optimal speed and accuracy of object detection. arXiv preprint. https://arxiv.org/abs/2004.10934

Bowes, B. D., Tavakoli, A., Wang, C., Heydarian, A., Behl, M., Beling, P. A., & Goodall, J. L. (2021) Flood mitigation in coastal urban catchments using real-time stormwater infrastructure control and reinforcement learning. Journal of Hydroinformatics, 23(3), p.pp. 529-547. https://doi.org/10.2166/hydro.2020.080

Center for western weather and water extremes (CW3E) (2021) CW3E Event Summary: 10-16 November 2021, Accessed at 12/14/2021 from https://cw3e.ucsd.edu/cw3e-event-summary-10-16-november-2021/

Chaudhary, P., D'Aronco, S., Moy de Vitry, M., Leitão, J. P., & Wegner, J. D. (2019) Flood-water level estimation from social media images. ISPRS Annals of the Photogrammetry, Remote Sensing and Spatial Information Sciences, 4(2/W5), p.pp. 5-12. https://doi.org/10.3929/ethz-b-000351581

Cohen, S., Raney, A., Munasinghe, D., & Loftis, J. D. (2019). The floodwater depth estimation tool (FwDET v2. 0) for improved remote sensing analysis of coastal flooding. Natural Hazards and Earth System Sciences, 19(9), p.pp. 2053. https://doi.org/10.5194/nhess-19-2053-2019

Cross, C., Farhadmanesh, M., & Rashidi, A. (2020) Assessing close-range photogrammetry as an alternative for lidar technology at UDOT divisions No. UT-20.18. Utah. Dept. of Transportation. Division of Research. https://rosap.ntl.bts.gov/view/dot/54923

Fan, Q., Yang, J., Hua, G., Chen, B., & Wipf, D. (2017) A generic deep architecture for single image reflection removal and image smoothing. In Proceedings of the IEEE International Conference on Computer Vision, p.pp. 3238-3247. https://doi.org/10.48550/arXiv.1708.03474

Federal Emergency Management Agency (FEMA) (2007) Floodplain management: principles and current practices. Academic Emergency Management and Related Courses (AEMRC) for the Higher Education Program. https://training.fema.gov/hiedu/aemrc/courses/coursetreat/fm.aspx

Federal Highway Administration (2004) Manual on uniform traffic control devices (MUTCD): standard highway signs. Accessed at 08/03/2021 from https://mutcd.fhwa.dot.gov/ser-shs_millennium_eng.htm

Fetanat, M., Stevens, M., Jain, P., Hayward, C., Meijering, E., & Lovell, N. H. (2021) Fully elman neural network: a novel deep recurrent neural network optimized by an improved Harris Hawks algorithm for classification of pulmonary arterial wedge pressure. IEEE Transactions on Biomedical Engineering. https://doi.org/10.1109/TBME.2021.3129459

Forati, A. M., & Ghose, R. (2021) Examining community vulnerabilities through multi-scale geospatial analysis of social media activity during Hurricane Irma. International Journal of Disaster Risk Reduction, p.pp. 102701. https://doi.org/10.1016/j.ijdrr.2021.102701

Fu, C. Y., Liu, W., Ranga, A., Tyagi, A., & Berg, A. C. (2017) DSSD: Deconvolutional single shot detector. arXiv preprint. https://arxiv.org/abs/1701.06659

Gebrehiwot, L. Hashemi-Beni, G. Thompson, P., & Kordjamshidi, T.E. (2019) Langan deep convolutional neural network for flood extent mapping using unmanned aerial vehicles data. Sensors, 19 (7), p.pp. 1486. https://doi.org/10.3390/s19071486

Han, D., Liu, Q., & Fan, W. (2018) A new image classification method using CNN transfer learning and web data augmentation. Expert Systems with

Applications, 95, p.pp. 43-56. https://doi.org/10.1016/j.eswa.2017.11.028

Han, Z., & Sharif, H. O. (2020) Vehicle-related flood fatalities in Texas, 1959–2019. Water, 12(10), p.pp. 2884. https://doi.org/10.3390/w12102884

Hao, W., & Zhili, S. (2020) Improved mosaic: algorithms for more complex images. In Journal of Physics: Conference Series, 1684(1), p.pp. 012094. IOP Publishing. https://iopscience.iop.org/article/10.1088/1742-6596/1684/1/012094/pdf

He, K., Zhang, X., Ren, S., & Sun, J. (2015) Spatial pyramid pooling in deep convolutional networks for visual recognition. IEEE Transactions on Pattern Analysis and Machine Intelligence, 37(9), p.pp. 1904-1916. https://doi.org/10.1109/TPAMI.2015.2389824

Hu, R., Zhang, S., Wang, P., Xu, G., Wang, D., & Qian, Y. (2020) The identification of corn leaf diseases based on transfer learning and data augmentation. In Proceedings of the 2020 3rd International Conference on Computer Science and Software Engineering, p.pp. 58-65. https://doi.org/10.1145/3403746.3403905

Hu, T. Y., Armandpour, M., Shrivastava, A., Chang, J. H. R., Koppula, H., & Tuzel, O. (2021) Synt++: Utilizing imperfect synthetic data to improve speech recognition. arXiv preprint, https://arxiv.org/abs/2110.11479

Insurance Bureau of Canada (2021) British Columbia floods cause $450 million in insured damage, Accessed at 12/09/2021 from http://www.ibc.ca/bc/resources/media-centre/media-releases/british-columbia-floods-cause-450-million-in-insured-damage

Kamari, M., & Ham, Y., (2021) Vision-based volumetric measurements via deep learning-based point cloud segmentation for material management in jobsites. Automation in Construction, 121, p.pp. 103430. https://doi.org/10.1016/j.autcon.2020.103430

Kingma, D. P., & Ba, J. (2014) Adam: A method for stochastic optimization. arXiv preprint https://arxiv.org/abs/1412.6980

Kong, N., Tai, Y. W., & Shin, J. S. (2013) A physically-based approach to reflection separation: from physical modeling to constrained optimization. IEEE Transactions on Pattern Analysis and Machine Intelligence, 36(2), p.pp. 209-221. https://doi.org/10.1109/TPAMI.2013.45

Lin, T. Y., Dollár, P., Girshick, R., He, K., Hariharan, B., & Belongie, S. (2017) Feature pyramid networks for object detection. In Proceedings of the IEEE Conference on Computer Vision and Pattern Recognition, p.pp. 2117-2125. https://arxiv.org/abs/1612.03144

Lin, T. Y., Maire, M., Belongie, S., Hays, J., Perona, P., Ramanan, D., Dollár, P., & Zitnick, C. L. (2014) Microsoft COCO: Common objects in context. In European Conference on Computer Vision. Springer, Cham, p.pp. 740-755. https://doi.org/10.1007/978-3-319-10602-1_48

Liu S, Qi L, Qin H, Shi J, & Jia J. (2018) Path aggregation network for instance segmentation. Proceedings of the IEEE Computer Society Conference on Computer Vision and Pattern Recognition, p.pp. 8759–68. https://arxiv.org/abs/1803.01534v4

Liu, W., Anguelov, D., Erhan, D., Szegedy, C., Reed, S., Fu, C. Y., & Berg, A. C. (2016) SSD: Single shot multibox detector. In European Conference on Computer Vision. Springer, Cham, p.pp. 21-37. https://doi.org/10.1007/978-3-319-46448-02

Lo, S. W., Wu, J. H., Lin, F. P., & Hsu, C. H. (2015) Visual sensing for urban flood monitoring. Sensors, 15(8), p.pp. 20006-20029. https://doi.org/10.3390/s150820006

Lyons, M. B., Keith, D. A., Phinn, S. R., Mason, T. J., & Elith, J. (2018) A comparison of resampling methods for remote sensing classification and accuracy assessment. Remote Sensing of Environment, 208, p.pp. 145-153. https://doi.org/10.1016/j.rse.2018.02.026

Mishra, A., Mukherjee, S., Merz, B., Singh, V. P., Wright, D. B., Villarini, G., Paul, S., Kumar, D. N., Khedun, C. P., Niyogi, D., Schumann, G., & Stedinger, J. R. (2022) An overview of flood concepts, challenges, and future directions. Journal of Hydrologic Engineering, 27(6), p.pp. 03122001. https://doi.org/10.1061/(ASCE)HE.1943-5584.0002164

Nath, N. D., & Behzadan, A. H., (2020) Deep convolutional networks for construction object detection under different visual conditions. Frontiers in Built Environment, 6, p.pp. 97. https://doi.org/10.3389/fbuil.2020.00097

National Oceanic and Atmospheric Administration (NOAA) (2021) Northwest River forecast center (NWRFC). Accessed at 12/14/2021 from https://www.nwrfc.noaa.gov/rfc/

Park, S., Baek, F., Sohn, J., & Kim, H. (2021) Computer vision–based estimation of flood depth in flooded-vehicle images. Journal of Computing in Civil Engineering, 35(2), p.pp. 04020072. https://doi.org/10.1061/(ASCE)CP.1943-5487.0000956

Redmon, J., Divvala, S., Girshick, R., & Farhadi, A. (2016) You only look once: Unified, real-time object detection. In Proceedings of the IEEE Conference on Computer Vision and Pattern Recognition, p.pp. 779-788. https://www.cv-foundation.org/openaccess/content_cvpr_2016/html/Redmon_You_Only_Look_CVPR_2016_paper.html

Ruder, S. (2016) An overview of gradient descent optimization algorithms. arXiv preprint.

https://doi.org/10.48550/arXiv.1609.04747

Sarel, B., & Irani, M. (2004) Separating transparent layers through layer information exchange. In European Conference on Computer Vision. Springer, Berlin, Heidelberg, p.pp. 328-341. https://doi.org/10.1007/978-3-540-24673-2_27

Shaghaghian, Z., & Yan, W. (2019) Application of deep learning in generating desired design options: experiments using synthetic training dataset. arXiv preprint. https://arxiv.org/abs/2001.05849

Sharif, H. O., Hossain, M. M., Jackson, T., & Bin-Shafique, S. (2012) Person-place-time analysis of vehicle fatalities caused by flash floods in Texas. Geomatics, Natural Hazards and Risk, 3(4), p.pp. 311-323.
https://doi.org/10.1080/19475705.2011.615343

Sharif, H. O., Jackson, T. L., Hossain, M. M., & Zane, D. (2015) Analysis of flood fatalities in Texas. Natural Hazards Review, 16(1), p.pp. 04014016.
https://doi.org/10.1061/(ASCE)NH.1527-6996.0000145

Sharma, K., Aggarwal, A., Singhania, T., Gupta, D., & Khanna, A. (2019) Hiding data in images using cryptography and deep neural network. arXiv preprint. https://doi.org/10.48550/arXiv.1912.10413

Smith, A., (2021) 2020 U.S. billion-dollar weather and climate disasters in historical context. Beyond the Data. Climate news, stories, images, & video. Accessed at 08/08/2021 from https://www.climate.gov/news-features/blogs/beyond-data/2020-us-billion-dollar-weather-and-climate-disasters-historical

Stone, W. C., Cheok, G., & Lipman, R. (2000) Automated earthmoving status determination. In Robotics 2000, p.pp. 111-119. https://doi.org/10.1061/40476(299)14

Tammina, S. (2019) Transfer learning using VGG-16 with deep convolutional neural network for classifying images. International Journal of Scientific and Research Publications (IJSRP), 9(10), p.pp. 143-150.
http://dx.doi.org/10.29322/IJSRP.9.10.2019.p9420

Tremblay, J., Prakash, A., Acuna, D., Brophy, M., Jampani, V., Anil, C., To, T., Cameracci, E., Boochoon, S., & Birchfield, S. (2018) Training deep networks with synthetic data: Bridging the reality gap by domain randomization. In Proceedings of the IEEE Conference on Computer Vision and Pattern Recognition Workshops, p.pp. 969-977. https://arxiv.org/abs/1804.06516v3

Turpin, A., & Scholer, F. (2006) User performance versus precision measures for simple search tasks. In 29th Annual International ACM SIGIR Conference on Research and Development in Information Retrieval, Seattle, WA, p.pp. 11–18. https://doi.org/10.1145/1148170.1148176

Wan, R., Shi, B., Hwee, T. A., & Kot, A. C. (2016) Depth of field guided reflection removal. In 2016 IEEE International Conference on Image Processing (ICIP) IEEE, p.pp. 21-25. https://doi.org/10.1109/ICIP.2016.7532311

Wang, C. Y., Liao, H. Y. M., Wu, Y. H., Chen, P. Y., Hsieh, J. W., & Yeh, I. H. (2020) CSPNet: A new backbone that can enhance learning capability of CNN. In Proceedings of the IEEE/CVF Conference on Computer Vision and Pattern Recognition Workshops, p.pp. 390-391. https://arxiv.org/abs/1911.11929

Ward, P. J., Van Pelt, S. C., De Keizer, O., Aerts, J. C. J. H., Beersma, J. J., Van den Hurk, B. J. J. M., & Te Linde, A. H. (2014) Including climate change projections in probabilistic flood risk assessment. Journal of Flood Risk Management, 7(2), p.pp. 141-151. https://doi.org/10.1111/jfr3.12029

Zhan, C., Ghaderibaneh, M., Sahu P., & Gupta, H. (2021) DeepMTL: Deep learning based multiple transmitter localization, In 2021 IEEE 22nd International Symposium on a World of Wireless, Mobile and Multimedia Networks (WoWMoM), p.pp. 41-50, https://doi.org/10.1109/WoWMoM51794.2021.00017

Zhu, L., Xie, Z., Liu, L., Tao, B., & Tao, W. (2021) IoU-uniform R-CNN: Breaking through the limitations of RPN. Pattern Recognition, 112, p.pp. 107816. https://doi.org/10.1016/j.patcog.2021.107816